# Recommendations for Comprehensive and Independent Evaluation of Machine Learning-Based Earth System Models

**Paul A. Ullrich**, Lawrence Livermore National Laboratory and University of California Davis
**Elizabeth A. Barnes**, Colorado State University
**William D. Collins**, Lawrence Berkeley National Laboratory
**Katherine Dagon**, NSF National Center for Atmospheric Research
**Shiheng Duan**, Lawrence Livermore National Laboratory
**Joshua Elms**, Indiana University Bloomington
**Jiwoo Lee**, Lawrence Livermore National Laboratory
**Ruby Leung**, Pacific Northwest National Laboratory
**Dan Lu**, Oak Ridge National Laboratory
**Maria J. Molina**, University of Maryland
**Travis A. O'Brien**, Indiana University Bloomington and Lawrence Berkeley National Laboratory
**Finn O. Rebassoo**, Lawrence Livermore National Laboratory

## Abstract

Machine learning (ML) is a revolutionary technology with demonstrable applications across multiple disciplines. Within the Earth science community, ML has been most visible for weather forecasting, producing forecasts that rival modern physics-based models. Given the importance of deepening our understanding and improving predictions of the Earth system on all time scales, efforts are now underway to develop forecasting models into Earth-system models (ESMs), capable of representing all components of the coupled Earth system (or their aggregated behavior) and their response to external changes. Modeling the Earth system is a much more difficult problem than weather forecasting, not least because the model must represent the alternate (e.g., future) coupled states of the system for which there are no historical observations. Given that the physical principles that enable predictions about the response of the Earth system are often not explicitly coded in these ML-based models, demonstrating the credibility of ML-based ESMs thus requires us to build evidence of their consistency with the physical system. To this end, this paper puts forward five recommendations to enhance comprehensive, standardized, and independent evaluation of ML-based ESMs to strengthen their credibility and promote their wider use.

## 1 Introduction

Earth system models (ESMs) have long been and continue to be invaluable tools for deepening our understanding of the Earth system and its response to forcing, such as greenhouse gasses, aerosols, land-use land-cover change, and solar irradiance. Scientists use ESMs to understand how the Earth system looked in the past and how it may look in the future. A wide array of end-users apply the data produced by ESMs to assess risk and inform adaptation planning. Despite their broad importance, traditional ESMs are also computationally expensive and incorporate sometimes crude assumptions about fundamental physical processes, leading to significant biases in key climatological fields. However, recent and rapid progress in machine learning (ML) has the potential to advance solutions to these problems. Building upon the successes of ML in other fields, ML-based forecast models have emerged that produce skillful weather forecasts at a fraction of the cost of conventional physics-based models (e.g., Lam et al., 2023; Ben Bouallègue et al., 2024). These efforts have stimulated recent work on ESMs that incorporate ML components or are entirely data-driven, including a variety of hybrid models, emulators, and foundation models (see Table 1 for a definition of these terms) (Eyring et al., 2024). Despite promising advancements, building trust in ML-based ESMs for scientific discovery or applications remains an outstanding challenge. **To this end, this paper outlines a plan for comprehensive, standardized, and independent evaluation of ML-based ESMs to strengthen their credibility for wider application, based on five core recommendations.**

ESMs are widely used for various scientific and applied purposes (see Table 2 for a summary of common applications). What sets ESMs apart from weather models is their capacity to perform experiments out-of-sample: ESMs need to be able to simulate the coupled Earth system under forcings outside the range of the period of instrumental record (generally since industrialization). This is primarily accomplished through adherence to the physical laws that govern the Earth system, including fundamental concepts such as conservation of mass, momentum, and energy, and consistency with emergent behavior such as geostrophic balance or constraints on globally averaged precipitation. If these laws are not explicitly enforced in the physics-based or ML-based model, it is essential to demonstrate that the respective model somehow exhibits them. With only one period of record covering a short and recent span of the Earth's history available to us, we need confidence that ESMs can do more than align with these observations; they need to get the right answer for the right reasons to be useful for understanding and projecting climate. Coming up with a testing strategy that goes beyond historical comparison requires us to build upon a long history of ESM evaluation and devise creative strategies to evaluate models' physical consistency.

Despite the challenges still ahead, ML-based ESMs are potentially a step-change in our ability to do meaningful science and deliver actionable information to decision-makers. Most remarkably, emulators and foundation models (see Table 1) provide an order-of-magnitude speed-up (and lower computing power per simulated year) compared with physics-based models. This enables the construction of very large ensembles that deliver improved uncertainty quantification (e.g., Mahesh et al., 2024a; Mahesh et al., 2024b; Kochkov et al., 2024), higher signal-to-noise ratios (relevant for detection and attribution), more exhaustive exploration of low-likelihood high-impact events and the tails of statistical distributions, and the capacity to generate future climate scenarios by interpolating between available ones.

Many in the climate community are rightly skeptical of the ability of emulators to generalize beyond the bounds of their training datasets: especially when changes in forcings are involved. Much work still remains to assess whether these models are credible for producing future projections, particularly under forcings that are far from recent history. A related question has been a central theme in ESM research over the past several decades: how do we assess the trustworthiness of an ESM's ability to operate in regimes where we do not know the correct answer? In the past, climate scientists have approached this question, in part, by various intercomparison efforts and hierarchical model evaluation approaches. We argue that such an approach is relevant for ML-based ESMs as well. This scientifically grounded effort to establish model credibility will allow ML-based ESMs to complement and expand upon the ecosystem of tools for understanding the Earth's past and future.

Motivated by the considerable excitement that has brewed around ML-based ESMs, this paper puts forward the following recommendations to the growing ML-based ESM development community for comprehensive and independent evaluation of these models:

1. Build upon the experience of the ESM community and its long history with model intercomparison while leveraging shared model code, data, and diagnostics.
2. Assemble a suite of idealized and simplified tests to evaluate the basic behavior of ML models.
3. Understand the behavior of ML-based ESMs in the ecosystem of Earth system modeling tools.
4. Develop an extensible evaluation framework that is widely accepted by the scientific community and relevant users.
5. Identify a trusted independent party to manage regular intercomparison of ML-based and physics-based models.

The remainder of this paper walks through these recommendations, connecting with the substantial work already done in ESM evaluation. It concludes with our draft proposal for a common evaluation strategy among ML-based ESMs, aiming to inspire further developments in this space.

**Table 1:** Descriptors commonly used to describe ML-based weather and climate models. Note that these descriptors are not meant to be mutually exclusive.

| **Descriptor** | **Summary** |
|---|---|
| Forecasting | Models trained primarily on historical weather data or operational forecast archives; these systems may neglect the coupling of Earth system components or climate-relevant fields altogether, like sea-surface temperatures or irradiance, which are essential for ESMs. |
| Autoregressive / Roll outs | Models whose output from the current iteration becomes the input for the next iteration. |
| Surrogate | Models trained to predict a specific quantity (e.g., global temperature change as a function of greenhouse gas concentration). |
| Hybrid | Traditional physics-based weather/climate model infrastructure with one or more parameterizations/components replaced with ML. |
| Emulator | Models trained to emulate a single physics-based weather/climate model. |
| Foundation | Models pre-trained on multiple, large datasets (e.g., CMIP6 models, reanalysis data products) that can be fine-tuned to support many downstream applications. |

**Table 2:** Applications of ESMs and potential roles of ML-based ESMs.

| ESM Application | Role of ML-Based ESMs |
|---|---|
| Predictions of weather, subseasonal to seasonal (S2S), or subseasonal to decadal (S2D) time scales | Rapid generation of huge forecast ensembles to quantify the probability of different climate states on weather/S2S/S2D time scales. ML-based models could be more skillful than physics-based models because of direct training on observations. |
| Assessing the risk of extreme weather or low-likelihood high-impact events | Rapid generation of huge forecast or projection ensembles to better quantify the frequency and character of rare events. |
| Detection and attribution | Cheap generation of large climate ensembles to increase signal-to-noise ratio; using explainable AI techniques to decompose temperature or precipitation time series into components of forcing (e.g., Sweeney and Fu, 2024). |
| Transient climate response and equilibrium climate sensitivity | Quantifying global temperature as a result of $CO_2$ and/or other radiative forcing and Earth system feedbacks across climate timescales. |
| Process understanding and hypothesis testing | Rapid testing of ideas before running more expensive physics-based model simulations; use of auto differentiation to probe input/output relationships. |
| Climate change projections and uncertainty quantification | Rapid generation of huge ensembles of future climate projections and associated uncertainties; computationally efficient quantifying of the model prediction uncertainty. |
| Assessment of climate impacts | Computationally efficient downscaling of coarse resolution climate projections to a finer resolution for assessment of regional climate impact. |
| Tipping points and "what if" science | Unclear. This topic is far out of sample, and so may require physics-based modeling at a fundamental level. |

## 2 Building on the Experience of the Earth System Modeling Community

Coordinated ESM evaluation has a long and storied history, originating in the early 1980s with Robert Cess' organization of the Feedback Analysis of GCMs (general circulation models) and Intercomparison with Observations (FANGIO) project (Potter et al., 2011). This effort, using prescribed sea surface temperature experiments, led to the first multi-model constraints on climate sensitivity (which have taken decades to refine; Sherwood et al. 2020), and was the basis for the Atmospheric Model Intercomparison Project (AMIP; Gates et al., 1992) and subsequent Coupled Model Intercomparison Projects (CMIPs; Meehl et al., 2000, 2005, Taylor et al., 2012, Eyring et al., 2016b). The consensus from this work was that intercomparison was effective at making better climate models (Meehl et al., 1997), with a steady increase in the fidelity of “top tier” models (e.g., Eyring et al., 2021, Ahn et al., 2022, Lee et al., 2023). A wealth of products have emerged from these past efforts, including model codes, data, and diagnostics, all of which need to be considered in the ML model landscape.

There is no doubt that the ML model development community should leverage these past efforts and draw upon the deep experience of the ESM evaluation community. Perhaps the lowest hanging fruit is through direct application of the many ESM evaluation toolkits now available. Observing systems have recorded key climate variables since the late 1800s, and have been supplemented over the past four decades by comprehensive remotely sensed measurements from an array of global satellites. The summary statistics derived from these observations are an essential reference for ESM performance and have been widely vetted for use in a wide variety of model benchmarking toolkits. Benchmarking toolkits incorporate suites of metrics and diagnostics that measure and depict consistency between model simulations and observations. Metrics and diagnostics evaluate meteorological patterns, mean climatology, climate variability (including well-known modes of variability such as the El Niño-Southern Oscillation), extreme weather events, physical processes, emergent relationships, and historical trends (under changes in observed forcing). A list of some more commonly employed packages is provided in Table 3; further information on available tools in the community can be found in Hoffman et al. (2024). Use of these evaluation tools would allow ML model developers to draw on decades of evaluation experience with physics-based ESMs, and feedback from the ML community would allow developers to augment their toolkits with benchmarks appropriate for ML models. These tools are also applied with relative ease; Figure 1 depicts a portrait plot from the PCMDI Metrics Package which shows evaluation results for the Ai2 Climate Emulator (ACE) as trained on the Energy Exascale Earth System Model (E3SM) v2, relative to other CMIP6 models. If such tools are widely employed in the ML model development community, intercomparison would be far easier between such models. Notably, Table 4 lists several efforts underway to develop evaluation packages targeted specifically at ML-based ESMs, though these are in the early stages.

Although the application of existing ESM benchmarking packages is an essential first step, there are unavoidable challenges and limitations in this approach. The majority of the metrics and diagnostics from these packages focus on comparing against historical observations, despite the historical record only covering a period of relatively stationary climate and only one possible meteorological trajectory. For many ML-based climate models, the need to avoid testing models against the data used to train them further limits our capacity to compare them against observations (in case those observations were used as training data). Just as physics-based models may be tuned to reduce model-observation differences, ML-based models may be trained to statistically reproduce the observations without learning the underlying physics or being able to extrapolate beyond the historical training period. This concern underscores the need for more creativity in devising evaluation strategies, some of which we explore here.

**Table 3:** Common benchmarking toolsets for evaluation of global ESMs.

| Tools | Primary Evaluation Focus | Citation |
|---|---|---|
| PCMDI Metrics Package (PMP) | Mean climate, variability modes, extremes | Lee et al., 2024 |
| International Land Model Benchmarks (ILAMB) | Land surface | Collier et al., 2018 |
| International Ocean Model Benchmarks (IOMB) | Ocean | Fu et al., 2022 |
| Cyclone Metrics Package (CyMEP) | Tropical cyclones | Zarzycki et al., 2021 |
| ESMValTool | Mean climate (and others) | Eyring et al., 2016a |
| Climate Variability Diagnostics Package for Large Ensembles (CVDP-LE) | Climate variability in large ensembles | Philips et al., 2020 |
| Model Diagnostics Task Force (MDTF) | Process-oriented diagnostics | Neelin et al., 2023 |

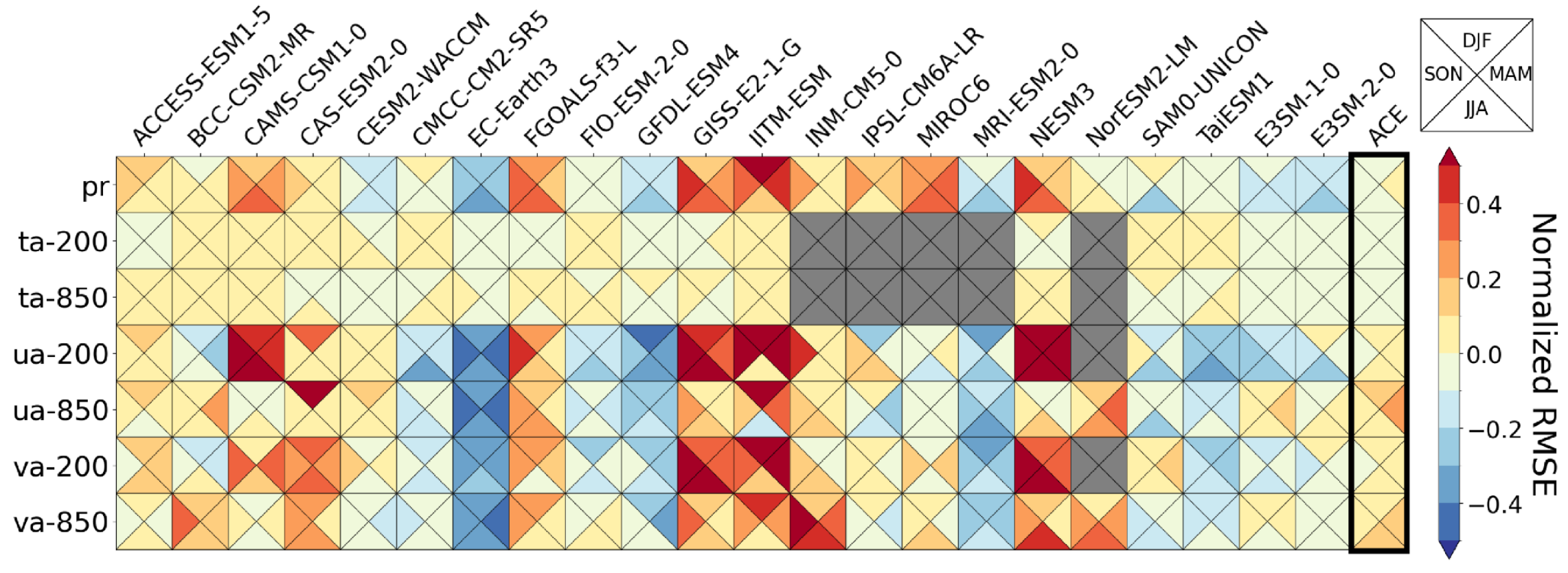

**Figure 1:** Tools like the PCMDI metrics package (PMP) can be easily adapted to evaluate ML-based climate modeling systems. Here the Ai2 Climate Emulator (ACE) trained on E3SMv2 is intercompared with CMIP6 models on seasonal measures of normalized root-mean-square error of precipitation (pr) compared to GPCP (Adler et al., 2018), along with ta-200 (200mb temperature), ta-850 (850mb temperature), ua-200 (200mb zonal wind speed), ua-850 (850mb zonal wind speed), va-200 (200mb meridional wind speed) and va-850 (850mb meridional wind speed) compared to ERA5 reanalysis data (Hersbach et al., 2020). The RMSE of each variable is normalized by the median RMSE of all models. The blue color indicates the model performs relatively better than other models, and the red color indicates the opposite. A result of +0.2 indicates an error that is 20 % greater than the median RMSE across all models, whereas -0.2 indicates an error that is 20% less than the median.

**Table 4:** Benchmarking toolsets for evaluation of global ML-based ESMs.

| Tools | Primary Evaluation Focus | Citation |
|---|---|---|
| Earth-2 MIP | Weather and climate | NVIDIA, 2024 |
| ClimateBench | Physically-based ESM emulators | Watson-Parris et al., 2022 |
| ClimDetect | Climate change detection and attribution | Yu et al., 2024 |
| ClimSim-Online | Hybrid ML-physics models | Yu et al., 2023 |
| WeatherBench 2 | Weather | Rasp et al., 2024 |

## 3 Beyond Historical Evaluation

Out-of-sample projections of the climate system are generally considered credible if they are grounded in physical principles (i.e., the conservation laws that underlie all of modern physics) and their emergent properties are consistent with our understanding of the Earth system. Conservation laws in the atmosphere were discussed at length in Thuburn (2008), and it was concluded that only a few of the conserved quantities could be used effectively to measure model performance. Indeed, beyond mass (and to a lesser degree energy, with appropriate tuning, and only in coupled ESMs), no other quantity is explicitly conserved in most physics-based ESMs. With that said, mass conservation lends itself to more than global metrics of invariance, it also means that tracer concentrations can't go negative: e.g., in each grid cell precipitation needs to be strictly limited by the amount of water vapor in the column. So these metrics provide a good start for evaluating models beyond their ability to reproduce the historical records. That said, these constraints are insufficient on their own: For instance, the conservation of mass and energy does not translate to any sort of useful constraint on equilibrium climate sensitivity, one of the most important quantities in climate science. Instead, some of the most scientifically and societally-relevant quantities emerge from a complex series of processes and feedbacks.

Evaluation of model performance can also rely on the properties that emerge from fundamental physics. From basic atmospheric science, we know air parcel saturation limits the amount of water vapor present in a grid cell before condensation and precipitation will occur; while supersaturation can occur in the upper troposphere, it is extremely rare for it to occur in the near-surface or mid-troposphere. More complex constraints could also be included. For instance, theory indicates that globally averaged precipitation increases at a rate ~1-2%/K while atmospheric water vapor increases at ~7%/K (Held and Soden, 2006; Pendergrass and Hartmann, 2014). Many other emergent constraints have also been explored in the literature in recent years, developed both as a means of assessing models' consistency with the real world and as an alternate pathway to link historical observations with future change (Hall et al., 2019; Williamson et al., 2021; Shiogama et al., 2022; Klein and Hall, 2015).

Power spectra have been used to reveal that some ML models tend to damp higher wavenumbers more strongly than physics-based models (Kochkov et al., 2024; Bonavita, 2024). These results suggest that the "effective resolution" of some ML-based weather and climate models is likely more coarse than the grid spacing might indicate, and is a likely product of the tendency for ML methods to regress to the mean. However, there are some recent examples where model architectural choices seem to avoid excessive damping (e.g. Mahesh et al. 2024a,b). Thus, metrics related to power spectra can help track effective versus grid resolution as new emulators are developed (Rasp et al. 2024).

Relationships between variables (i.e., covariances) provide another mechanism to test consistency within the model. For example, heat waves are often associated with weaker wind speeds since they are largely associated with high-pressure ridges. Recent work by Zhang and Boos (2023) also highlighted the temporal pattern in precipitation that occurs around the hottest days of the year, where dry conditions are generally present before temperature peaks, followed by moist conditions once high temperatures drive sufficient atmospheric instability to initiate convection. Such behavior could readily inform the development of an associated metric or diagnostic.

Feature structure is another avenue for model evaluation where ML-based forecasting models have struggled (Bonavita, 2024). Feature tracking packages such as TempestExtremes (Ullrich et al., 2021) and the Toolkit for Extreme Climate Analysis (TECA; Loring et al., 2023) can be employed to track tropical cyclones, extratropical cyclones, monsoon depressions, atmospheric rivers, and other important atmospheric features. Features present in other Earth system components, such as marine heatwaves or sea ice extent, could also be incorporated. Composites of the environmental structure can then reveal anomalous or unphysical behavior.

## 4 Revisiting Simplified Tests in the Age of Machine Learning

Given the inherent complexity of physics-based ESMs, evaluation begins far before most physical processes are even incorporated into the model. Software engineering operates in much the same way – rather than only testing the software package once all the functionality has been implemented, unit testing is employed to validate each function as it is added. Applied to an ESM, hierarchical evaluation means focusing first on the simplest model that exhibits behavior consistent with the component of the Earth system being tested (e.g., dry dynamics with no topography or surface fluxes for the atmosphere), making sure physical laws are obeyed at each timestep, and gradually adding complexity. This process allows us to evaluate ESM behavior in a simpler context, where analytic or semi-analytic solutions are available. Notably, O'Loughlin et al. (2024) advocate for such a modular design as a way to develop a component-level understanding of ML models.

In the case of the atmospheric component model, testing begins with the dynamical core, which is responsible for solving the fluid flow equations in the thin spherical annulus that wraps the Earth. Major physical components are then added one at a time, beginning with moisture and moist processes, ocean-atmosphere fluxes, topography, and finally a land surface and fully coupled ocean (see below). Such a traditional ESM evaluation hierarchy for the atmospheric model component is given in Table 5.

In a manner similar to the atmospheric component model, testing of the land component can begin with the biophysical core, which is responsible for simulating terrestrial water and energy balance processes. Major physical components are then added one at a time, beginning with vegetation dynamics and photosynthesis, carbon and nitrogen cycles, topography-driven runoff, and finally human land use and land cover change (Fisher and Koven 2020). Throughout this process, the high heterogeneity of land surfaces necessitates testing across diverse landscapes to ensure robust performance across varied environmental conditions (Byrne et al. 2024). When coupling to the atmosphere, an important consideration for land modeling is the representation of surface turbulent fluxes, and notably sub-daily fluxes, for capturing land-atmosphere feedbacks (Chaney et al. 2024).

Simplified tests for the ocean component can follow the recommendations of Bishnu et al. (2024), which outlines a suite of shallow water test cases with increasing complexity. Additional examples of simplified tests for the coupled system include testing the ML atmosphere model coupled to a hierarchy of ocean models such as slab ocean, mixed layer ocean, and full ocean models (Hsu et al. 2022).

**Table 5:** Traditional ESM evaluation hierarchy (atmospheric model component).

| Step in the Hierarchy | What is Tested |
|---|---|
| 2D shallow water experiments (e.g., Williamson et al., 1992) | Horizontal fluid dynamics on a sphere, barotropic energy exchange across scales |
| 3D without moisture (e.g., Jablonowski and Williamson, 2006) | Baroclinic dynamics (e.g., baroclinic instability, one of the most fundamental atmospheric processes), baroclinic energy exchange across scales |
| 3D with moisture and simplified physics (e.g., Reed and Jablonowski, 2012) | Isolated dynamical processes in the global system |
| Aquaplanet experiments (e.g., Hess et al., 1993) | Atmospheric model behavior without land-atmosphere or dynamic ocean |
| Fixed sea-surface temperature runs (e.g., Gates et al., 1999) | Atmosphere and land-surface behavior with tight constraints on model drift |
| Fully coupled simulations (e.g., Gates et al., 1995) | Fully coupled modeling system |

For ML-based ESMs, such a hierarchy may seem unnecessary if the underlying model framework is kept relatively simple. However, these tests are important to demonstrate that the fundamental physical laws are respected in the model in such a way that important emergent phenomena spontaneously manifest (e.g., baroclinic Rossby waves in the atmosphere). They exercise a specific model's ability to simulate physics using initial states that are well outside the model's training dataset: effectively building trust in the models by demonstrating (or failing to demonstrate) generalizability.

With that said, the traditional ESM evaluation hierarchy does not necessarily translate to ML-based models: especially tests that involve activating/deactivating/swapping specific model components that represent specific components of the Earth system. ML-based models often lack explicitly distinct components representing specific physical processes, unlike their physics-based counterparts. Consequently, deactivating particular physics (e.g., moist physics) in these ML models may not be practical through simple component deactivation. Instead, it might require retraining the entire model using data from a physics-based model where the specific process has been deactivated. However, retraining a ML-based model for a particular test (e.g., as needed for an atmosphere model running the 2D shallow water experiments) introduces new challenges with interpreting what can be learned about the more general ML-based ESM. Several of the idealized atmospheric test cases highlighted in Table 5 can be run in certain ML-based ESMs, assuming that those models allow for prescribed topography and radiative forcing. For instance, the baroclinic instability test of Jablonowski and Williamson (2006) is a dry dynamical core test case with prescribed topography and zero radiative forcing, with a well-documented numerical solution for short-period runs, suitable for a forecasting model or ESM. This test has been essential in evaluating atmospheric dynamical cores, given the importance of baroclinic instability to global weather patterns. Similarly, the tropical cyclone test of Reed and Jablonowski (2012) could be run in any ESM with at least 0.5° grid spacing, as long as topography can be removed entirely, and would be helpful for assessing the ability of a model to capture the structure of simulated tropical cyclones. Even advection could be tested under the formulation documented in the 2008 Dynamical Core Model Intercomparison Project (DCMIP) test case document for test 1-0-y (Jablonowski et al, 2008). Other such tests are documented as part of the 2008, 2012, and 2016 DCMIP test case documents (Jablonowski et al, 2008; Ullrich et al., 2012; Ullrich et al., 2016).

To demonstrate that ML-based models are capable of performing such idealized tests, despite these tests being outside the range of their training data, we show results from a moist baroclinic wave test using the Spherical Fourier Neural Operators (SFNO) model of Bonev et al. (2023). We implement the initial condition specification of Bouvier et al. (2023), which uses an analytical formula to define the initial conditions. This choice was made because other baroclinic wave tests require numerically integrating quantities in a way consistent with a model’s dynamical core: this is not possible in most emulators since the numerical discretization scheme is not explicitly knowable. Results from this test are depicted in Figure 2. While the model does produce a Rossby wave, the test immediately reveals an important, albeit expected, deviation of SFNO from the behavior of physics-based models: the imprint of topography and continental geometry immediately becomes evident in many fields, despite the initial condition being zonally uniform. This expected behavior emerges because SFNO internalizes the impact of Earth’s topography and land-sea contrast on atmospheric flow, since all of the training data involved real Earth topography. To the best of our knowledge, no presently available ML-based ESM supports arbitrary, prescribed topography. Nonetheless the test remains stable and the results otherwise consistent with expectations, highlighting the power in utilizing idealized tests to evaluate ML-based models.

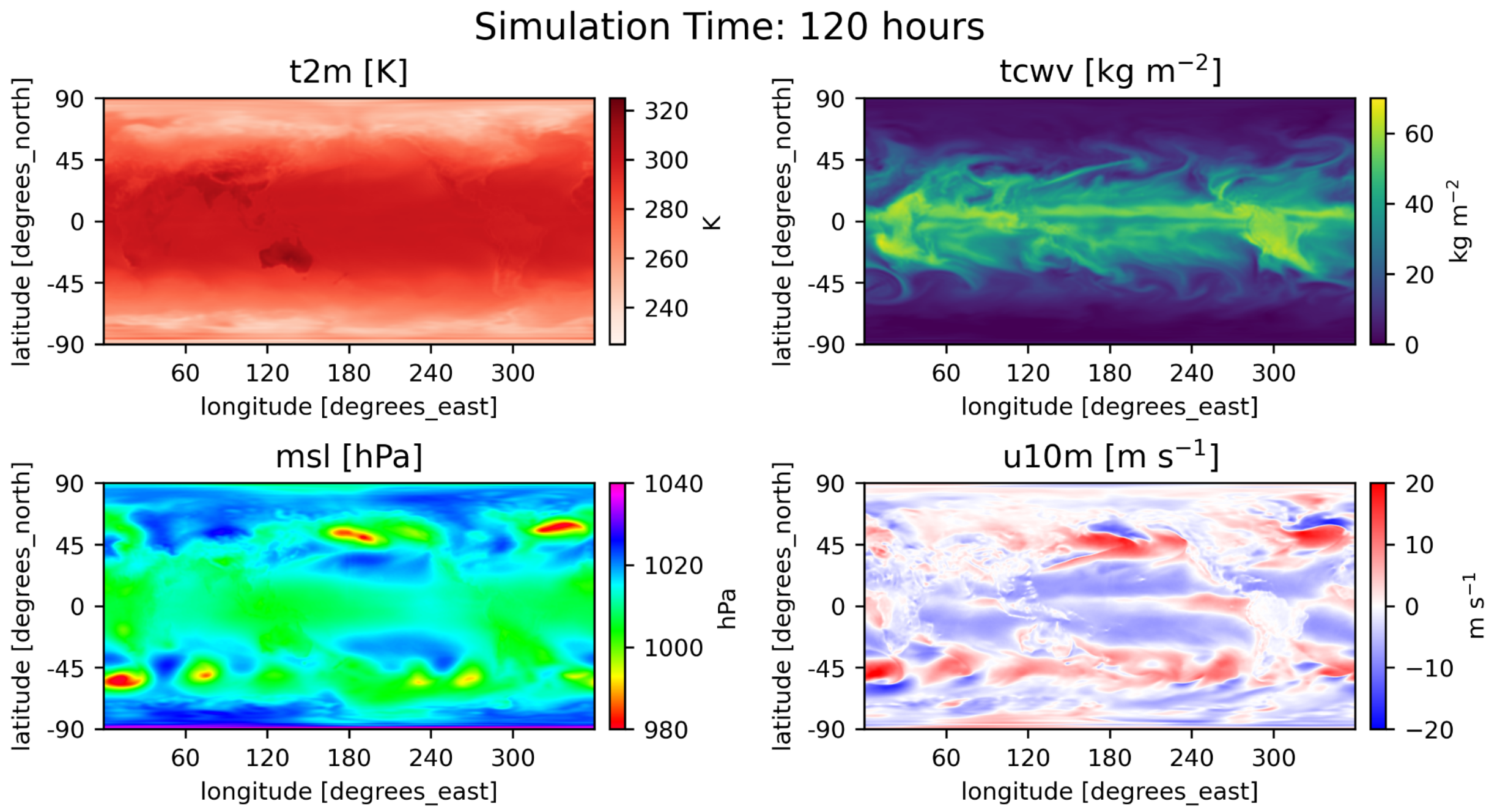

**Figure 2:** Snapshots of 2m temperature (t2m), total column water vapor (tcwv), mean sea-level pressure (msl) and 10m wind speed (u10m) from the Bouvier et al. (2023) idealized baroclinic wave test at 120h after initialization, as simulated using the SFNO model of Bonev et al. (2023).

Besides those tests commonly employed in physics-based model development, there are also tests that are unnecessary in most physics-based ESMs since the explicit numerical methods used by these ESMs encode local causality directly. However, in ML models, causality can and should be explicitly tested. To do so one could begin with a near-steady atmospheric flow and add a large local perturbation at the scale of a single grid point. The propagation of that disturbance should be limited to the acoustic wave (in the case of a hydrodynamic perturbation) or the maximum wind speed (in the case of a tracer perturbation). Such a test could reveal if the ML model allows information to propagate faster than physics would allow.

In recent work by Hakim and Masanam (2024), four idealized tests were conducted with realistic topography and a climatological mean steady state, which may be more suitable for ML-based models that do not allow modification of topography. These tests include steady tropical heating with Matsuno–Gill response, an extratropical cyclone test analogous to Jablonowski and Williamson (2006) except over the Pacific, a test of geostrophic adjustment, and a test looking at Atlantic tropical cyclogenesis. The authors used these novel experiments to conclude that Pangu-Weather (an ML-based global forecasting model; Bi et al., 2022) does likely encode physical principles since its behavior is consistent with physical understanding even for tests in which the idealized initial conditions are far outside the distribution of initial conditions in the training data.

## 5 ML Models are Part of the Broader Ecosystem

While we may not be able to definitively conclude whether ML-based ESMs are producing the right answer for the right reasons, we can evaluate if the ML-based models are within or outside the envelope of pre-existing models under alternate forcings (e.g., under projected warming). Namely, we can compare projections from ML-based models side-by-side with existing physics-based models, reduced complexity models (Nicholls et al., 2021), and statistical/dynamical downscaling methods. The toolkits discussed in section 2 could be used for these experiments, except instead of comparing models against observations, the alternate forcing experiment would be compared to the respective historical simulation. This approach could then be used to cover a variety of variables, regions, and experiments. One such effort in this direction is ClimateBench (Watson-Parris et al., 2022), which uses training data from the Norwegian Earth System Model (NorESM2; Seland et al., 2020) and measures models' consistency with these projections. However, a large enough ensemble of ESM projections would be needed to delineate internal variability from forced response (e.g., Lütjens et al. 2024).

Of course, care needs to be taken in case a discrepancy arises between the ML models and the other physics-based models in the ecosystem. One should not simply assume that the ML models are behaving inconsistently with physical principles, as the reasons for these differences could be far more nuanced and can include physics-based ESM biases.

Hybrid models can also be paired with their "sibling" (i.e., the ESM used to embed ML) and emulators paired with their "parent" (i.e., the physics-based ESM(s) on which they are trained), providing another route for examining consistency. Evaluations could include pairwise hybrid-to-sibling or emulator-to-parent comparisons against observational metrics, allowing us to identify similarities or differences. Alternatively, in the context of a multi-model ensemble (MME), is the projection spread of hybrid models or emulators distinguishable from the MME of traditional physics-based ESMs? These comparisons open opportunities to reveal and understand the behaviors of emulators, hybrid models, and physics-based models beyond benchmarking using direct comparison with observations.

## 6 Developing a Framework for Intercomparison

There is no doubt that the significant successes in ESM development over the past four decades can be attributed to regular intercomparison and healthy competition and collaboration among modeling groups. However, frameworks for intercomparison require agreement and investment from all involved. Using our past experience with evaluating CMIP class ESMs as guidance, there is a clear need for all of the following:

- common metrics and diagnostics recipes that all groups can agree upon to measure success,

- commitment to open source code, pre-trained ML-based ESMs, data access, user documentation for each experiment, and model/data provenance from modeling groups, to ensure transparency, fairness, and reproducible science,
- a consistent output format (for when data is written to disk), compatible with widely accepted, Climate and Forecasting (CF) metadata convention standards (Hassell et al., 2017; Eaton et al., 2022, http://cfconventions.org/), and
- an independent organization responsible for validating and collating results from development groups.

The aforementioned framework would be effective at building trust in these ML-based ESM models, and making it clear that nothing is being hidden from data users. The objective would be to provide a buffer against model development groups cherry-picking metrics that show good performance, ensure that all models are evaluated on a level playing field, and support intercomparison by the broad community of ESM developers, climate scientists, and users of ESM model outputs.

With a common, prescribed metrics recipe, there are always concerns that groups will engage in "metric hacking" – namely, training the ML-based model or tuning the physics-based model to maximize their performance scores as a shortcut  to building a skillful model. However, we would also argue that there should not be a single metric to summarize performance (it is not even clear if any single metric would make sense for something as complicated as an ESM). If groups are obliged to show all metrics, then optimizing among all metrics becomes nearly impossible even with sophisticated uncertainty quantification, multi-objective optimization, and model calibration methods. Regardless, such optimization would simply result in a model that is more consistent with physical principles, if these evaluation metrics are well defined. Inevitably, it is necessary to pick some quantitative measure of skill (i.e., a loss function) for training an ML system, and efforts to devise better loss functions that enable more comprehensive performance gains could be an exciting and innovative direction for research.

## 7 Summary

Although ML-based ESMs are in their infancy, there is no better time than now to lay out common standards for ML-based ESM evaluation and encourage common targets for intercomparison. Physics-based ESMs have been undergoing standardized intercomparison for forty years, and in that time have made considerable progress in their ability to simulate the Earth system with minimal intervention. By leveraging the knowledge, experience, and capabilities built up by the scientific and model evaluation community over this time, comprehensive evaluation of ML-based ESMs is achievable in a relatively short time. Our discussion of benchmarks for this new generation of models is summarized in Table 6 – by no means intending to be the final say on the types of evaluation that should be performed, but a starting point for a conversation between the development and evaluation communities. The authors of this paper have already begun work to codify this menu, yielding greater insight into the capability of ML-based models to simulate the Earth system out-of-sample.

**Table 6:** An example of a menu for ML-based ESM evaluation.

<table>
<tr>
<td>
<b>Benchmarking Packages (§2)</b><br>
PCMDI Metrics Package (PMP)<br>
● Mean climate<br>
● ENSO metrics<br>
● Tropical and extratropical modes of variability<br>
● Monsoon metrics<br>
● Extremes metrics<br>
● Precipitation variability<br>
● Cloud feedbacks[a]<br>
International Land Model Benchmarks (ILAMB)[b]<br>
International Ocean Model Benchmarks (IOMB)[c]<br>
Cyclone Metrics Package (CyMEP)<br>
ESMValTool<br>
Model Diagnostics Task Force (MDTF)<br>
<br>
<b>Sanity checks (§3)</b><br>
Global mass conservation<br>
Non-negative tracer mass<br>
No near-surface supersaturation<br>
<br>
<b>Performance metrics (§3)</b><br>
Zonal means<br>
Power spectra<br>
Covariances<br>
Feature (e.g., TC) frequency, character and structure<br>
Historic (1900-present) temperature trends under historic forcing
</td>
<td>
<b>Emergent constraints (§3)</b><br>
Global water vapor change (~7%/K)<br>
Global precipitation change (~1-2%/K)<br>
Ratio of OLR to Niño-3.4 index<br>
Others?<br>
<br>
<b>Idealized Test Cases (§4)</b><br>
Steady-state advection (DCMIP2008)[d]<br>
Baroclinic wave test case (DCMIP2008)[d]<br>
Mountain-induced Rossby wave (DCMIP2008)[d]<br>
3D Rossby-Haurwitz wave (DCMIP2008)[d]<br>
Moist baroclinic instability (DCMIP2012)[d]<br>
Tropical cyclone test (DCMIP2012)[d]<br>
Causality test<br>
Steady tropical heating (H&M2024)<br>
Extratropical cyclone development (H&M2024)<br>
Geostrophic adjustment (H&M2024)<br>
Atlantic hurricane development (H&M2024)
</td>
</tr>
<tr>
<td colspan="2">
[a] AMIP and AMIP +4K experiments required<br>
[b] Land model evaluation<br>
[c] Ocean model evaluation<br>
[d] Only models that allow prescription of bottom topography
</td>
</tr>
</table>

## Acknowledgments

The AMIP-type ACE simulation shown in Figure 1 was funded by LLNL-LDRD Project 22-ERD-052 PlusUp. We thank C. Bonfils, G. Pallotta and P. Caldwell for associated discussions. Works of LLNL-affiliated authors were performed under the auspices of the U.S. DOE by the Lawrence Livermore National Laboratory (LLNL) (contract no. DE-AC52-07NA27344) and their efforts were supported by the Regional and Global Model Analysis (RGMA) program area of the U.S. Department of Energy (DOE) Office of Science (SC), Biological and Environmental Research (BER) program. PNNL is operated for DOE by Battelle Memorial Institute under contract DE-AC05-76RL01830. Oak Ridge National Laboratory is operated by UT Battelle, LLC, for the DOE under contract DE-AC05-00OR22725. LBNL's contributions were supported by the Director, Office of Science, Office of Biological and Environmental Research of the U.S. Department of Energy under Contract No. DE-AC02-05CH11231. This material is based upon work supported by the U.S. Department of Energy, Office of Science, Office of Biological & Environmental Research (BER), Regional and Global Model Analysis (RGMA) component of the Earth and Environmental System Modeling Program under Award Number DE-SC0022070 and National Science Foundation (NSF) IA 1947282. This work was also supported by the National Center for Atmospheric Research (NCAR), which is a major facility sponsored by the NSF under Cooperative Agreement No. 1852977. This work was also supported, in part, by DOE BER as part of the Program for Climate Model Diagnosis and Intercomparison Project. This research was also supported in part by the Environmental Resilience Institute, funded by Indiana University's Prepared for Environmental Change Grand Challenge initiative and in part by Lilly Endowment, Inc., through its support for the Indiana University Pervasive Technology Institute. This work is also supported by DOE BER RGMA under University Award Number DE-SC0024093. LLNL-JRNL-870866-DRAFT.